\documentclass[letterpaper, 10 pt, conference]{ieeeconf}  

\IEEEoverridecommandlockouts   

\makeatletter
\let\NAT@parse\undefined
\makeatother

\usepackage[hidelinks]{hyperref}
\usepackage{url} 

\usepackage{comment} 
\usepackage{graphics} 
\usepackage{epsfig} 
\usepackage{amsmath}
\usepackage{amssymb}
\usepackage{siunitx}
\usepackage[official]{eurosym}
\usepackage{csquotes}
\usepackage[capitalise]{cleveref}

\let\labelindent\relax
\usepackage{enumitem}
\usepackage{booktabs}

\usepackage{multirow}
\usepackage[table]{xcolor}

\definecolor{mygray}{gray}{.95}

\usepackage{placeins}

\usepackage{subcaption}
\DeclareCaptionFormat{twolinetab}{#1#2#3}
\usepackage{tabularx}
\usepackage{graphicx}
\usepackage{adjustbox}
\usepackage{array}
\usepackage{placeins}
\usepackage{diagbox}

\usepackage[style=ieee, url=false, maxbibnames=3, doi=false, isbn=false, eprint=true]{biblatex}
\DeclareFieldFormat[misc]{title}{\mkbibquote{#1}}

\usepackage{makecell}
\usepackage{fontawesome}

\newcommand{\metrabs}[0]{MeTRAbs}

\renewcommand{\arraystretch}{0.85} 

\usepackage{pifont}
\newcommand{\cmark}{\textcolor{green}{\ding{51}}} 
\newcommand{\xmark}{\textcolor{red}{\ding{55}}}   

\usepackage[acronym]{glossaries}
\newacronym{vit}{ViT}{Vision Transformer}
\newacronym{cnn}{CNN}{Convolutional Neural Network}
\newacronym{gcn}{GCN}{Graph Convolution Network}

\newacronym{mlp}{MLP}{Multi-layer Perceptron}

\newacronym{jepa}{V-JEPA}{Video Joint-Embedding Predictive Architecture}
\newacronym{metrabs}{MeTRAbs}{Metric-Scale Truncation-Robust Heatmaps for Absolute 3D Human Pose Estimation}
\newacronym{hdgcn}{HD-GCN}{Hierarchically Decomposed Graph Convolution Network}
\newacronym{clip}{CLIP}{Contrastive Language-Image Pretraining}

\newacronym{ava}{AVA}{Atomic Visual Actions}
\newacronym{har}{HAR}{Human Action Recognition}

\newacronym{vfm}{VFM}{Video Foundation Model}
\newacronym{vlm}{VLM}{Video-Language Model}
\newacronym{hpe}{HPE}{Human Pose Estimation}
\newacronym{fps}{FPS}{Frames Per Second}

\newacronym{llm}{LLM}{Large Language Model}

\newacronym{ema}{EMA}{Exponential Moving Average}
\newacronym{gelu}{GELU}{Gaussian Error Linear Unit}

\newacronym{roi}{RoI}{Region of Interest}
\newacronym{mcvqa}{MC-VQA}{Multiple-Choice Video Question Answering}

\usepackage{listings}

\usepackage{afterpage}

\title{\LARGE \bf How do Foundation Models Compare to Skeleton-Based Approaches \\
 for Gesture Recognition in Human-Robot Interaction?}

\author{Stephanie Käs$^{1}$,
        Anton Burenko$^{1}$,
        Louis Markert$^{1}$,  \\
        Onur Alp \c{C}ulha$^{1}$,
        Dennis Mack$^{2}$, 
        Timm Linder$^{2}$, Bastian Leibe$^{1}$%
        \thanks{$^{1}$ Chair for Computer Vision, RWTH Aachen University, Germany. Mail: 
                \textrm{\{kaes, leibe\}@vision.rwth-aachen.de}              
                \newline
                $^{2}$ Robert Bosch GmbH, Corporate Research \& Bosch Center for AI, Renningen and Hildesheim, Germany. Mail: 
                \textrm{\{firstname.lastname\}@de.bosch.com}
        }
} 
        
\begin{document}
\maketitle
\thispagestyle{empty}
\pagestyle{empty}

\begin{abstract}
Gestures enable non-verbal human-robot communication, especially in noisy environments like agile production. Traditional deep learning-based gesture recognition relies on task-specific architectures using images, videos, or skeletal pose estimates as input. Meanwhile, Vision Foundation Models (VFMs) and Vision Language Models (VLMs) with their strong generalization abilities offer potential to reduce system complexity by replacing dedicated task-specific modules. This study investigates adapting such models for dynamic, full-body gesture recognition, comparing V-JEPA (a state-of-the-art VFM), Gemini Flash 2.0 (a multimodal VLM), and HD-GCN (a top-performing skeleton-based approach). We introduce NUGGET, a dataset tailored for human-robot communication in intralogistics environments, to evaluate the different gesture recognition approaches. 
In our experiments, HD-GCN achieves best performance, but V-JEPA comes close with a simple, task-specific classification head---thus paving a possible way towards
reducing system complexity, by using it as a shared multi-task model. In contrast, Gemini struggles to differentiate gestures based solely on textual descriptions in the zero-shot setting, highlighting the need of further research on suitable input representations for gestures.
\end{abstract}

\section{Introduction}

Symbolic navigation gestures are widely used in aviation and traffic control for non-verbal communication \cite{data:natops:song,data:police:he, data:police-gcn:liu, data:traffic:wiederer}, especially in noisy environments. Similarly, mobile robots in warehouses and factories could benefit from robust gesture recognition for effective human robot interaction (HRI).
Most gesture recognition approaches rely on task-specific architectures. Image-/video-based methods often require dedicated backbones to encode visual features \cite{data:police:he, gesture:LSTM:mullick, gesture:HRGVit:tan, gesture:TransformerHand:eusanio}, while skeleton-based approaches depend on a human pose estimation (HPE) module running on the robot \cite{HPE:LSTM_gesture:wu, data:body-gesture-recognition:laplaza, data:gcn-for-gesture:guo, HPE:Metrabs:Sarandi}. Meanwhile, general-purpose Vision Foundation Models (VFMs) and Vision-Language Models (VLMs) are evolving rapidly \cite{Other:V-JEPA:Bardes, Other:Dinov2:Oquab, other:Gemini:2025, other:openai2025gpt4, video:surveyVFM:madan}. Trained on vast, internet-scale resources, they possess strong generalization capabilities and could potentially replace several perception components in robotics (e.g., human and object perception, task planning, navigation). They can be fine-tuned, their embeddings used as classifier inputs, or prompted for zero-shot tasks like multiple-choice classification and video understanding \cite{AR:VideoGLUE:yuan}. However, research on applying VFMs and VLMs for gesture recognition remains limited.

\begin{figure}[t]
  \centering
  \includegraphics[width=\columnwidth]{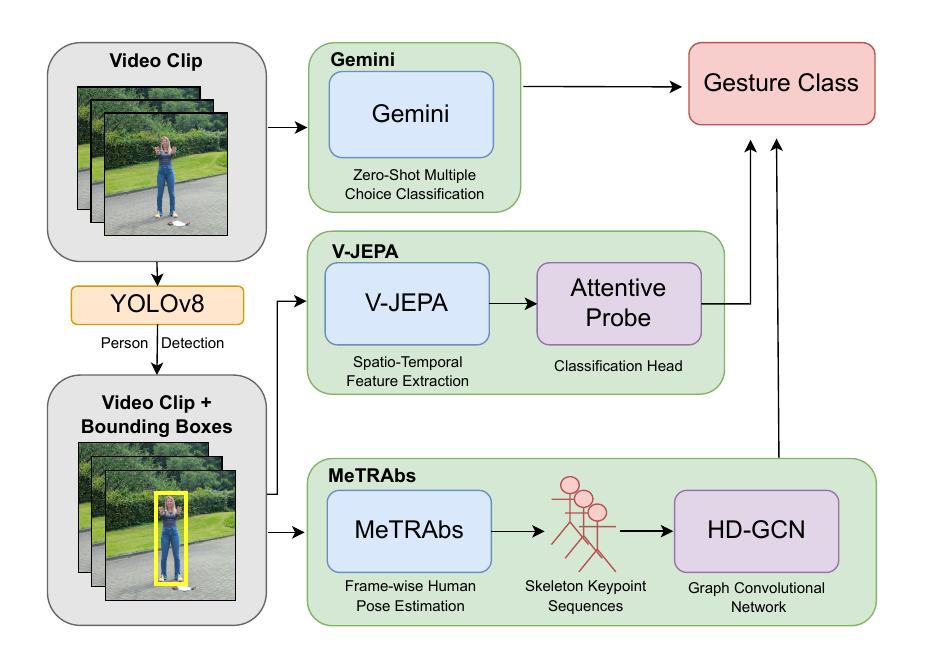}
  \caption{We compare Gemini Flash 2.0 and V-JEPA, two Vision Foundation Models, to HD-GCN, a skeleton-based method, on the task of full-body gesture classification using NUGGET, our novel dataset for hands-free human-robot communication. Our results (\cref{Experiments}) show that a VFM with a task-specific classification head approaches the accuracy of HD-GCN, while the VLM relying on prompt engineering in zero-shot mode performs significantly worse.}
  \label{fig:pipeline}
  \vspace*{1ex}
\end{figure}

In this work, we therefore want to explore the use of such foundation models (FM) for classifying dynamic HRI gestures. 
We use HD-GCN~\cite{AR:HD-GCN:Lee}, a graph convolutional network (GCN) for skeleton-based action recognition, as a baseline and compare it to two state-of-the-art FM approaches:
V-JEPA \cite{Other:V-JEPA:Bardes}---a video FM that we equip with a fine-tuned classification head for gesture recognition, and Gemini \cite{other:geminifamily:google}---a multimodal FM that we use as a zero-shot classifier. 

Prior work on gesture recognition mostly focuses on static or fine-grained hand gestures \cite{ HRI:hand_gesture_survey:mitra, HRI:hand_gesture_survey:linardakis, HRI:hand_gesture_survey:qi}, which are less feasible for mobile robotics as they are hard to interpret from low resolution images over long distances \cite{HRI_long_sistance_gestures:beeri}, or when the hands do not face the camera. 
We thus introduce and evaluate on NUGGET, a novel dataset for single- and multi-person dynamic gesture recognition in human-robot communication designed for intralogistic use cases. NUGGET emphasizes intuitive body gestures via arm movements, excluding fine-grained hand gestures and facial expressions, making it suitable for robots operating in noisy long-distance scenarios.

In summary, our contributions are:
1.) We comprehensively compare two FMs, V-JEPA and Gemini Flash 2.0, to the skeleton-based HD-GCN, assessing the potential of VFMs and VLMs for whole-body gesture recognition.
2.) We introduce NUGGET, a novel dataset for gesture-based robot navigation, emphasizing intuitive body language over hand and facial expressions.

\section{Related Work}

\subsection{Using Gestures in Human Robot Interaction}

Gesture recognition has diverse applications, spanning sign language, symbolic gestures in traffic, aviation and military (\cite{data:police:he, data:police-gcn:liu, data:traffic:wiederer, data:natops:song}), medical and mobile robotics or VR. Past works developed gestures specifically for human-machine interaction (\cite{data:uav:perera, data:auth-uav:patrona, data:rocogv2:Reddy}),  collaboration tasks~\cite{data:HRC:Chen, HRC:gesture:Xia, HRC:collab:ende} like person following~\cite{HRC:quadfollow:nasser}, robotic manipulation~\cite{HRI:manipulation:Cornak}, or manufacturing~\cite{HRC:gesture_review:Liu}. 
Our work specializes on \emph{upper-body gestures} for interaction with mobile intralogistics robots, where instructions like  ``go left'', ``slow down'' or ``lift the pallet'' are necessary. This multi-class classification task is closely related to full-body human action recognition (HAR) \cite{HAR:survey:sedaghati:2025, HAR:survey:poppe:2010} (``walking'', ``sitting'' etc.). Gesture recognition distinguishes \textit{static} and \textit{dynamic} approaches, where static recognition classifies still images with stationary gestures, and dynamic recognition incorporates temporal context from video or skeleton sequences \cite{HRI:dynamic_gesture_review:shi}. Our focus, as in \cite{data:body-gesture-recognition:laplaza}, is on recognizing \emph{dynamic} gestures from short temporal sequences.

\subsection{Technical Approaches for Gesture/Action Recognition}
Based on input type, vision-based gesture recognition divides into two categories: 1.) Skeleton-based approaches that process outputs of a preceding human pose estimator (HPE), 
and 2.) e.\,g. CNN- or Transformer-based ``end-to-end'' approaches that operate directly on image/video input.

 \subsubsection{\textbf{Skeleton-based Methods}}
 These methods use 2D/3D HPE  algorithms \cite{HPE:surveyMonocular2Dand3D:Liu} to extract skeletons
 or meshes. 
Commonly, resulting skeletons are processed using  graph-convolutional neural networks (GCNs) \cite{AR:HD-GCN:Lee, AR:TSGCNeXT:Liu, AR:Knowledge-Assisted:Nguyen} which capture spatio-temporal dependencies in structured representations. 
Further works use Autoencoder- or Transformer-based embedding representations \cite{AR:ViPLO:Park, AR:SkeleTR:Duan, AR:SkeletonMAE:Yan}
to transform skeletal data into a compact feature space prior to classification. Dynamic gesture tasks leverage temporal information e.g. via recurrent neural networks (RNNs/LSTMs) \cite{HPE:LSTM_gesture:wu}. Others incorporate semantic knowledge from LLMs \cite{AR:Knowledge-Assisted:Nguyen}.
Skeleton-based motion-to-text generation \cite{other:motiongpt:jiang} is still limited in capturing subtleties of
body language and gestures \cite{other:motiongpt2:wang}. 
We leverage HD-GCN \cite{AR:HD-GCN:Lee}, a hierarchical graph-convolutional method which addresses limitations of earlier graph-based learning approaches \cite{AR:Spatial-Temporal-GCN:Yan} that either relied too heavily on direct physical connections -- ignoring distant node relationships -- or indiscriminately aggregated all nodes, losing fine-grained spatial structure. To mitigate these issues, HD-GCN introduces a Hierarchically Decomposed Graph with center of mass nodes as structural anchors and an Attention-Guided Hierarchy Aggregation module, refining feature selection by emphasizing semantically relevant joint interactions via various pooling operations  \cite{AR:HD-GCN:Lee}. We use HD-GCN in combination with MeTRAbs~\cite{HPE:Metrabs:Sarandi,HPE:Autoencoder:Sarandi,Ours:fishnchips}, a 3D pose estimator.

Skeleton-based approaches benefit from HPE's abstraction of human appearance, reducing background noise in gesture classification as well as the need for large training sets compared to pure image-based gesture classifiers.

\definecolor{lightgray}{gray}{0.9}

\begin{table}[t]
    \centering
    \caption{Overview of datasets including static and dynamic full-body or upper-body communication gestures.}
    \label{tab:gesture:data}
    \vspace*{-2ex}
    \setlength{\tabcolsep}{4pt}
    \renewcommand{\arraystretch}{1.2}
    \label{tab:gesture_recognition_dss}
    \resizebox{\linewidth}{!}{
    \rowcolors{2}{white}{lightgray}
    \begin{tabular}{l| c c c c} 
    \hline
    \textbf{Dataset} & \textbf{Classes} & \textbf{Samples} & \textbf{Subjects} & \textbf{Public} \\
    \hline
    AUTH UAV Gesture \cite{data:auth-uav:patrona} & 6 & 4,930 & 8 & \xmark \\

    Body Gesture \cite{data:body-gesture-recognition:laplaza} & 15 & 150 & 10 & \xmark \\

    Body Hand Gesture \cite{data:gcn-for-gesture:guo} & 10 & 80 & 10 & \faEnvelopeO \\

    Chinese Traffic Police (CTPG) \cite{data:police:he} & 8 & 3354 & N/A & \cmark \\

    hiROS \cite{gestures:user_study:tan} & 27** & \textasciitilde7500 & 190 & \xmark \\

    IAS-Lab Collab. HAR \cite{data:ias-lab:terreran} & 10 (com.) & 540 & 6 & \xmark \\

    Keck Gesture \cite{data:keck:lin} & 14 & 294 & 3 & \cmark* \\

    Kinteract \cite{data:kinteract:borghi} & 10 & 168 & 10 & \cmark* \\

    KU Gesture \cite{data:ku:hwang} & 30 (com.) & \textasciitilde 3000 & 20 & \xmark \\

    NATOPS \cite{data:natops:song} & 24 & 9,600 & 20 & \cmark \\

    RoCoG-v2 (real) \cite{data:rocogv2:Reddy}  & 7 & 482 & 10 & \cmark \\

    RoCoG-v2 (synthetic) \cite{data:rocogv2:Reddy}  & 7  & 106,996 & N/A & \cmark \\

    Traffic Control (TCG) \cite{data:traffic:wiederer} & 4 & 250 & 5 & \faEnvelopeO \\

    Traffic Police Gestures \cite{data:police-gcn:liu} & 8 & 320 & 8 & \xmark \\

    Unnamed (Guo et al.) \cite{data:unnamed:fuad} & 5 & 450 & 5 & \xmark \\

    UAV-Gesture \cite{data:uav:perera} & 13 & \textasciitilde 700 & 8 & \faEnvelopeO \\
    
    \hline
    \textbf{NUGGET (ours)} & \textbf{15} &  \textbf{11,569} & \textbf{13} & \cmark \\
    \hline
    
    \end{tabular}}
    \smallskip
    \raggedright

    \textbf{Notation:} (com.) = Subset of communicative gestures, datasets also includes other actions. (**) = Partially requires fingers.\\
    (*) = Download link broken as of 03/2025, please check repositories for updates. (\faEnvelopeO) =  Upon request.
\end{table}

\subsubsection{\textbf{Image/Video-based Methods}}

These approaches classify gestures or actions directly from raw images. 
Therefore, they can incorporate additional scene context like fore- or background objects. 
Methods range from CNN-LSTM/RNN pipelines \cite{data:police:he, gesture:LSTM:mullick}, to Transformer-based models for action \cite{AR:SF-DETR:Kim, AR:Cross-Modal-Learning:Kim} and hand recognition~\cite{gesture:HRGVit:tan, gesture:TransformerHand:eusanio}.
A pre-VLM era survey \cite{FMg:zeroshot_action:estevam} examines zero-shot action recognition.

Large-scale FMs have recently achieved remarkable success in various tasks, being trained on massive datasets, enabling strong generalization. The VideoGLUE benchmark \cite{AR:VideoGLUE:yuan} (incl. action recognition) compares seven publicly available VFMs like DINO v2 \cite{Other:Dinov2:Oquab}, V-JEPA \cite{Other:V-JEPA:Bardes} using four adaptation methods: end-to-end fine-tuning, a frozen backbone, a frozen backbone with a multi-layer attention pooler, and a low-rank adapter.  Few-shot fine-tuning of FMs on limited data has proven effective in action recognition \cite{FMg:action_fewshot:vu, FMg:action_fewshot:wang}. 

VLMs enable \textbf{zero-shot classification}
by leveraging semantic knowledge and prompting techniques. Two main approaches exist: (a)~\textit{Text Comparison}—Multimodal models like GPT-4 \cite{other:openai2025gpt4}, Gemini \cite{other:Gemini:2025}, or specialized video FMs \cite{video:surveyVFM:madan} generate textual descriptions of video frames, that can be compared with ground truth (GT) text using cosine similarity or another LLM. Description detail and imperfect similarity metrics limit this method. (b)~\textit{Multiple Choice Prompting}—Here, class descriptions are provided in the prompt, and the model selects the best match.
Prior work on FM-based gesture recognition \cite{FMg:kNNroboco:reday} found training-free methods are limited  in quality (especially for synthetic videos), recommending fine-tuning after comparing kNN (ViT-B/16) to zero-shot ViCLIP on RoCoG-v2 \cite{data:rocogv2:Reddy}. However, as FMs rapidly evolve, we compare zero-shot prompting using SOTA-leader Gemini 2.0 \cite{other:Gemini:2025} against a VFM-based classifier (V-JEPA \cite{Other:V-JEPA:Bardes}) and a skeleton-based approach (MeTRAbs + HD-GCN \cite{AR:HD-GCN:Lee}) in our experiments.

\subsection{Whole-Body Gesture Recognition Datasets}

We observe that there is no established benchmark for upper-body or whole-body gesture recognition. Relevant datasets (\cref{tab:gesture:data}) include drone control \cite{data:uav:perera, data:auth-uav:patrona, data:rocogv2:Reddy} and standardized military gestures \cite{data:natops:song}.  Among them, the mostly synthetic RoCoG-v2 \cite{data:rocogv2:Reddy} 
and various traffic control datasets \cite{data:police:he, data:police-gcn:liu, data:traffic:wiederer} focus on directing vehicles (no manipulation gestures). Others feature general interactions like pointing \cite{data:ku:hwang}
or collaborative assembly gestures \cite{data:ias-lab:terreran}. 
Both \cite{gestures:user_study:tan} and \cite{data:body-gesture-recognition:laplaza} studied intuitive human-robot communication. 
In \cite{gestures:user_study:tan}, 190 participants were asked to perform various HRI gestures to achieve specified goals, but without being given detailed instructions on how the gestures should look like. The authors observed high inter-person variances which can hinder model training, and thus concluded that standardized gestures should be defined.  Related studies led to the creation of the \textit{Body Gesture} \cite{data:body-gesture-recognition:laplaza} dataset. As \cite{data:body-gesture-recognition:laplaza} is comparably small and \cite{gestures:user_study:tan} partially relies on finger motion and is not publicly available, we now present our novel, publicly available upper-body gesture dataset.

\section{A Novel Multi-Person Gesture Recognition Dataset for Robot Navigation}
\label{sec:NUGGET}
\begin{figure}[t]
  \centering
  \vspace*{-1ex}
    \centering

  \includegraphics[width=0.920\columnwidth]{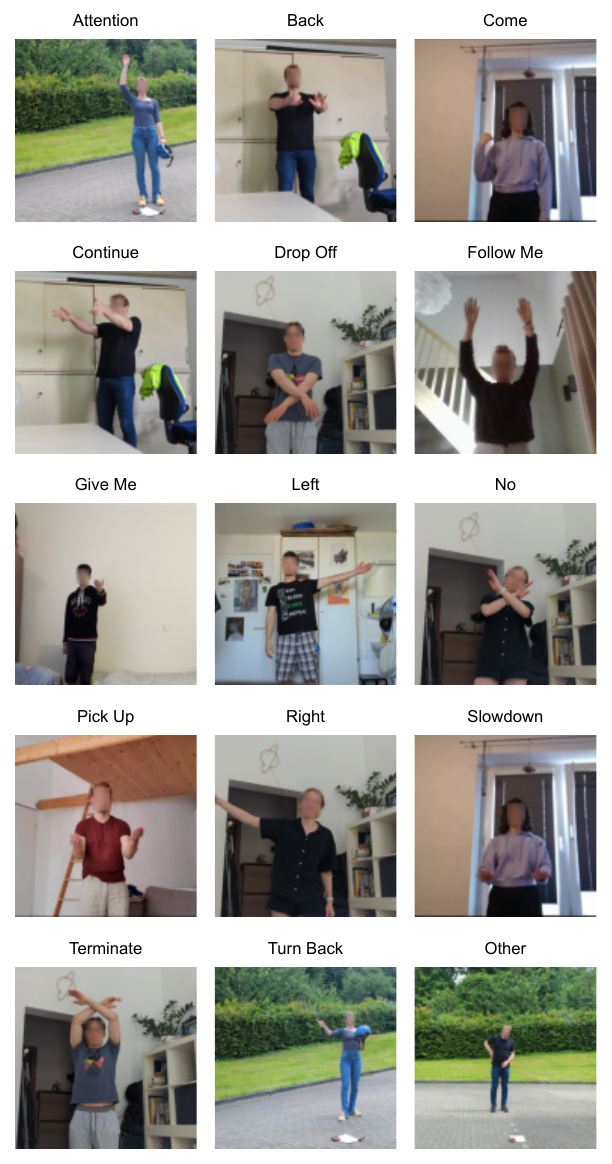}
  \caption{Photos of the 15 classes in our NUGGET dataset, comprising 14 intuitive upper-body HRI gestures and one miscellaneous class (\textit{Other}) for non-gesture inputs.}
  \label{fig:beispiel}

\end{figure}

We introduce \textbf{NUGGET} (\textit{\underline{N}atural \underline{U}ser \underline{G}estures for Robot \underline{G}uidance \underline{T}asks}), a dataset designed for multi-person gesture recognition in HRI tasks, specifically for intralogistics. It contains more than 11k pre-annotated sequences of 4\,s length, 1\,h of recordings (7.6\,GB) from 13 participants across 15 indoor/outdoor setups, 
capturing 14 gesture classes and a miscellaneous class (see \cref{tab:gestures}, \cref{fig:beispiel}). NUGGET includes  varying person-camera distances, lighting conditions, occlusions, object interactions and clothing styles. 

Our gestures are inspired by studies \cite{data:body-gesture-recognition:laplaza} and \cite{gestures:user_study:tan}, but specifically designed for intuitive interaction with (intralogistics) robots, without relying on fine-grained hand or finger poses that are difficult to detect at larger distances. Instead, we focus on arm-centric upper-body gestures. 
We include the following modifications to \cite{data:body-gesture-recognition:laplaza}, thereby adapting it for mobile robots in intralogistics scenarios: Remove \textit{Shrug}. Combine \textit{Random} and \textit{Static} into \textit{Other}. Combine \textit{No} and \textit{Stop}. Combine \textit{Yes} and \textit{Continue}. Add \textit{Follow me} (robot follows the person), \textit{Terminate} (robot stops tracking after perceiving \textit{Attention}). Add: \textit{Pick up} (robot lifts an object), \textit{Drop off} (robot releases an object), and \textit{Give me} (robot hands over object). Add \textit{Other} for non-gesture movements.

The data was recorded with various cameras (Galaxy S10, Google Pixel 6, HP Pavilion x360, Huawei P30, iPhone 12, Vivo V21e; 8-64 MP). Gesture labeling was semi-automated, with human validation to ensure accuracy. We create bounding box (BBox) sequences (tracks) with unique IDs for every person using the \textit{motpy} library \cite{Other:Motpy:Muron}, then manually assign start/stop timestamps and gesture class for each segment. As accurate temporal segmentation is notoriously difficult even for a human, labels may be slightly inconsistent at the individual frame level; however, our later evaluations are all based on full 4-second sequences.

\begin{figure}
    \begin{subfigure}{0.49\linewidth}
        \includegraphics[width=\textwidth]{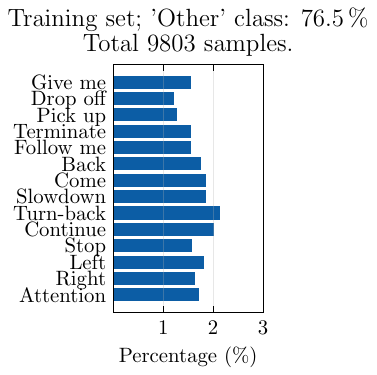}
        \label{fig:muiltitrain}
    \end{subfigure}
    \begin{subfigure}{0.455\linewidth}
        \includegraphics[width=\textwidth]{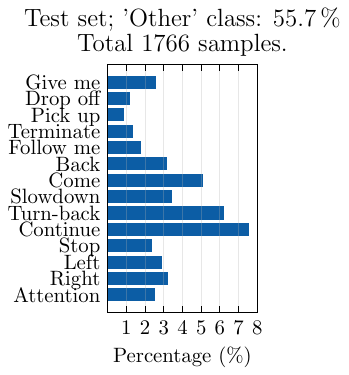}
        \label{fig:multitest}
    \end{subfigure}
    \vspace{-5ex} 
    \caption{Class distributions by percentage of samples per class in the NUGGET training and test set.}
    \label{fig:classdist}
\end{figure}

Unlike comparable datasets \cite{data:body-gesture-recognition:laplaza, gestures:user_study:tan, data:traffic:wiederer}, NUGGET videos contain up to three people in one frame. This way, we allow multi-person gesture recognition for future experiments on multi-user robot responses. Note that we use a single person test split in this paper (see \cref{tab:NUGGET_split}) as the Gemini VLM is used zero-shot without person detector in our main experiments.  As shown in \cref{fig:classdist}, gesture classes are more imbalanced in the test set to reflect real-world conditions. The inclusion of a high percentage of non-gesture activities ensures robustness against diverse human movements, e.\,g. natural moving of arms during human-human interactions.

\definecolor{lightgray}{gray}{0.9}

\begin{table}[t]
    \centering
    \setlength{\tabcolsep}{4pt}
    \renewcommand{\arraystretch}{1.1}
    \caption{Overview of all gesture classes in NUGGET.}
    \label{tab:gestures}
    \vspace*{-2ex}
    \resizebox{\linewidth}{!}{
    \begin{tabular}{llp{3.5cm}cc}
        \toprule
        \textbf{ID} & \textbf{Gesture} & \textbf{Movement} & \textbf{\#Arms} & \textbf{Dynamic} \\
        \midrule
        \rowcolor{lightgray}
        0  & Attention   & Hand up & 1 & \xmark \\
        1  & Right       & Right hand up on a side & 1 & \xmark \\
        \rowcolor{lightgray}
        2  & Left        & Left hand up on a side & 1 & \xmark \\
        3  & Terminate   & Cross hands above the head & 2 & \xmark \\
        \rowcolor{lightgray}
        4  & Follow me   & Two hands up & 2 & \xmark \\
        5  & Come        & Wave hand(s) towards oneself & 1--2 & \cmark \\
        \rowcolor{lightgray}
        6  & No/Stop        & Both arms crossed in front of the chest & 2 & \xmark \\
        7  & Drop off    & Cross hands at the bottom & 2 & \xmark \\
        \rowcolor{lightgray}
        8  & Yes/Continue    & Up/Down forearm motion in front of the body & 1--2 & \cmark \\
        9  & Turn back   & Circular movement with a single arm & 1 & \cmark \\
        \rowcolor{lightgray}
        10 & Slowdown    & Up/Down hand movement & 1--2 & \cmark \\
        11 & Back        & Wave hand(s) from oneself & 1--2 & \cmark \\
        \rowcolor{lightgray}
        12 & Pick up     & Lift both arms upwards & 2 & \cmark \\ 
        13 & Give me     & Hold straight hand in front & 1 & \xmark \\
        \rowcolor{lightgray}
        14 & Other       & Any other human presence &  &  \\
        \bottomrule
    \end{tabular}
    }
    \vspace*{2ex}
\end{table}

\definecolor{lightgray}{gray}{0.9}

\begin{table}[t] 
    \centering
    \renewcommand{\arraystretch}{1.1}
    \caption{Overview of how we split NUGGET data for our experiments in  \textbf{11.569 4\,s-long sequences} at 10\,FPS.} 
    \label{tab:NUGGET_split}
    \vspace*{-2ex}    
    \rowcolors{2}{white}{lightgray}
    \begin{tabularx}{\columnwidth}{l|X X X X}
    \toprule
      &   &  & \multicolumn{2}{c}{Persons/Frame~~~~~~~} \\
    \cline{4-5}
     \multirow{-2}{*}{NUGGET} & \multirow{-2}{*}{Gestures} & \multirow{-2}{*}{Other} & Single & Multi \\
    \hline
    Train & 9803 & 2306 &  7497 & 2260 \\
    Test & 1766 &  782 &   984  & 0 \\
    \bottomrule
    \end{tabularx}
\end{table}

\section{Methodology}

This study compares three algorithms for dynamic gesture recognition on our NUGGET dataset: HD-GCN, V-JEPA, and Gemini Flash 2.0. HD-GCN models spatial-temporal skeleton dynamics via graph structures, V-JEPA is a VFM that has learned rich representations through self-supervised learning, and the Gemini VLM can classify videos through prompt-based interaction. We now present details on how we adapt these methods to the dynamic gesture recognition task. 

\subsection{Skeleton-based Approach: \metrabs{} + HD-GCN}
For each video, we extract frames and apply a frozen body pose estimator—MeTRAbs—to generate joint-based skeleton representations. These per-frame skeletons are then associated over time \cite{Other:Motpy:Muron} and fed as skeleton sequences into HD-GCN \cite{AR:HD-GCN:Lee}, which has been trained from scratch on the gesture classes from the NUGGET train split. The following sections provide a detailed explanation of both components.

\paragraph{MeTRAbs}
MeTRAbs \cite{HPE:Metrabs:Sarandi} estimates absolute 3D human poses from image crops using known camera parameters. The model employs a EffNetV2-backbone to predict 2D and relative 3D joint coordinates, which are transformed into normalized image coordinates. A strong perspective model, solved via linear least squares~\cite{HPE:Metrabs:Sarandi}, then reconstructs the absolute 3D pose. In our pipeline, we use a frozen version of MeTRAbs trained on 28 datasets, as described in \cite{HPE:Autoencoder:Sarandi}. While in this work, we rely on standard pinhole cameras with a limited field of view (90--120°), MeTRAbs is also extendable to fisheye imagery \cite{Ours:fishnchips} to allow close-up interactions, or gesture commands when standing to the side of the robot.

\paragraph{Hierarchically Decomposed Graph Convolutional Network (HD-GCN)}  

To establish a gesture classification baseline independent of FMs, we required a robust state-of-the-art action recognition model capable of handling diverse and high-variance motion patterns. HD-GCN \cite{AR:HD-GCN:Lee} stood out due to its strong performance on several benchmarks like NTU RGB-D  \cite{data:ntu-rgb+d-120:liu} (120 action classes), with 90.1\% accuracy using fewer parameters than other top-performing models, such as DeGCN (91.0\%) \cite{AR:DeGCN:Myung}, LLM-assisted LA-GCN (90.7\%) \cite{AR:Knowledge-Assisted:Nguyen}, or TSGCNeXt (90.2\%) \cite{AR:TSGCNeXT:Liu}. This makes it particularly suited for computationally constrained applications like real-time gesture recognition on mobile robots.

\subsection{VFM + Classification Head: V-JEPA}
As a representative of non-language foundation models, we use V-JEPA (Video Joint Embedding Predictive Architecture) \cite{Other:V-JEPA:Bardes} with a frozen Vision Transformer ViT-L/16 \cite{Other:Image-Transformer:Dosovitskiy} backbone for feature extraction, and stack a self-trained gesture classifier on top. Unlike HD-GCN \cite{AR:HD-GCN:Lee}, which requires an intermediate pose estimation step, V-JEPA directly generates feature representations at the video level, bypassing the need for explicit keypoint detection.

 Similar to the skeleton-based approach, we adopt a top-down strategy, where a person detector (YOLOv8, see \cite{other:YOLO_surve:terven}) identifies regions of interest (RoI) in the video before applying V-JEPA for feature extraction. A task-specific classification head then uses these features for gesture recognition. 
 We evaluate two classification strategies: \textbf{Linear Probing} and (nonlinear) \textbf{Attentive Probing}.  
\textit{Linear Probe}: \label{sec:lincls}
 To obtain a compact video representation, the features are temporally averaged. Features within RoIs are then extracted using RoIAlign \cite{heMaskRCNN2018} and further pooled into a single feature vector. This vector is then fed into a linear classifier for gesture recognition. Linear probing uses either the \textit{last layer} or a concatenation of \textit{multiple layers} from the ViT \cite{bardesRevisitingFeaturePrediction2024}. 
 \textit{Attentive Probe}:
A cross-attention between the clip-level features and the pooled local features as the query is performed as in \cite{AR:VideoGLUE:yuan} to select the most relevant features from the final ViT layer for the gesture task. After this, the extracted features are combined with the query representation through a residual connection, normalized, and processed by a non-linear two-layer MLP before passing through a linear layer to predict the gesture class as in \cite{Other:V-JEPA:Bardes}. \\

\subsection{VLM + Zero-Shot Classification: Gemini Flash 2.0}
Vision-language FMs like GPT-4 Vision \cite{other:openai2025gpt4} or Gemini \cite{other:geminifamily:google} offer the chance to eliminate the need for manual data annotation or training as they can be used out-of-the-box on a prompt-basis for a variety of tasks. To classify gestures without task-specific training data, we use Gemini, a family of multimodal LLMs by Google DeepMind. Gemini models excel in zero-shot tasks as well as when finetuned on several benchmarks (like EgoSchema \cite{other:Gemini:2025}, VideoMME \cite{other:videomme_fu}) by leveraging extensive pretraining on diverse textual and visual data. Compared to GPT models, which outstand in text processing, Gemini specializes at handling large multimodal inputs and long-context understanding, due to its huge context window of up to 1\,mio. (Flash 2.0) to 2\,mio. (Pro 1.5) tokens \cite{other:Gemini:2025}.
Gemini employs a Mixture-of-Experts (MoE) architecture within a Transformer framework. It excels at complex video understanding through MoE’s specialized video processing experts.  We focus on applying Gemini Flash 2.0, which is designed for efficiency and low latency, balancing speed, explainability, and prediction quality \cite{other:Gemini:2025}. 
Instead of classifying individual frames in isolation, Gemini receives either MP4 videos or batches of consecutive frames (10 FPS) in our experiments, allowing it to incorporate temporal context when making predictions. This enables the VLM to recognize motion patterns, transitions, and dynamic characteristics of gestures without explicit sequence modeling. The model’s large context window ensures that even when multiple frames are analyzed together, it can effectively capture dependencies across time. We utilize prompts such as ``Given the following list of textual descriptions of gesture classes, decide which gesture class is most likely. Return only the class ID: $<$list$>$''. For more details on prompts, please refer to Sec.~\ref{Ablation}, ``Prompting Strategies for Gemini''.

\section{Experiments}
\label{Experiments}

\subsection{Overview}
The goal of our experiments is to evaluate the performance of selected state-of-the-art models\footnote{Model configurations are evaluated in ablation studies (\cref{Ablation}).} for full-body gesture recognition. 
We compare the following approaches on the test split of our NUGGET dataset, establishing a baseline for future research:
(1) \textbf{MeTRAbs + HD-GCN}, (2) \textbf{V-JEPA + Attentive Probing Classification Head}, (3) \textbf{Zero-Shot Multiple Choice Gemini Flash 2.0}.
We assess model performance using Jaccard Index, F1-Score, Top-1 and Top-3 Accuracy. Furthermore, we provide confusion matrices (with true positive rates, TPR) for a per-class performance overview and identification of challenging gestures, helping to highlight potential pitfalls for future downstream tasks. 

\subsection{Results}
\cref{tab:final_results} shows that MeTRAbs + HD-GCN outperforms V-JEPA by $\sim$4\%, with both achieving $>\!\!90\%$ accuracy. Gemini underperforms (42.1\% top-1) but reaches 79.2\% top-3 accuracy, indicating partial gesture understanding.

\begin{figure*}[h] 
    \centering
    \captionsetup[subfigure]{justification=centering,margin={0cm,0.4cm}}
    \begin{subfigure}[t]{0.32\textwidth}
        \centering
        \includegraphics[width=\linewidth]{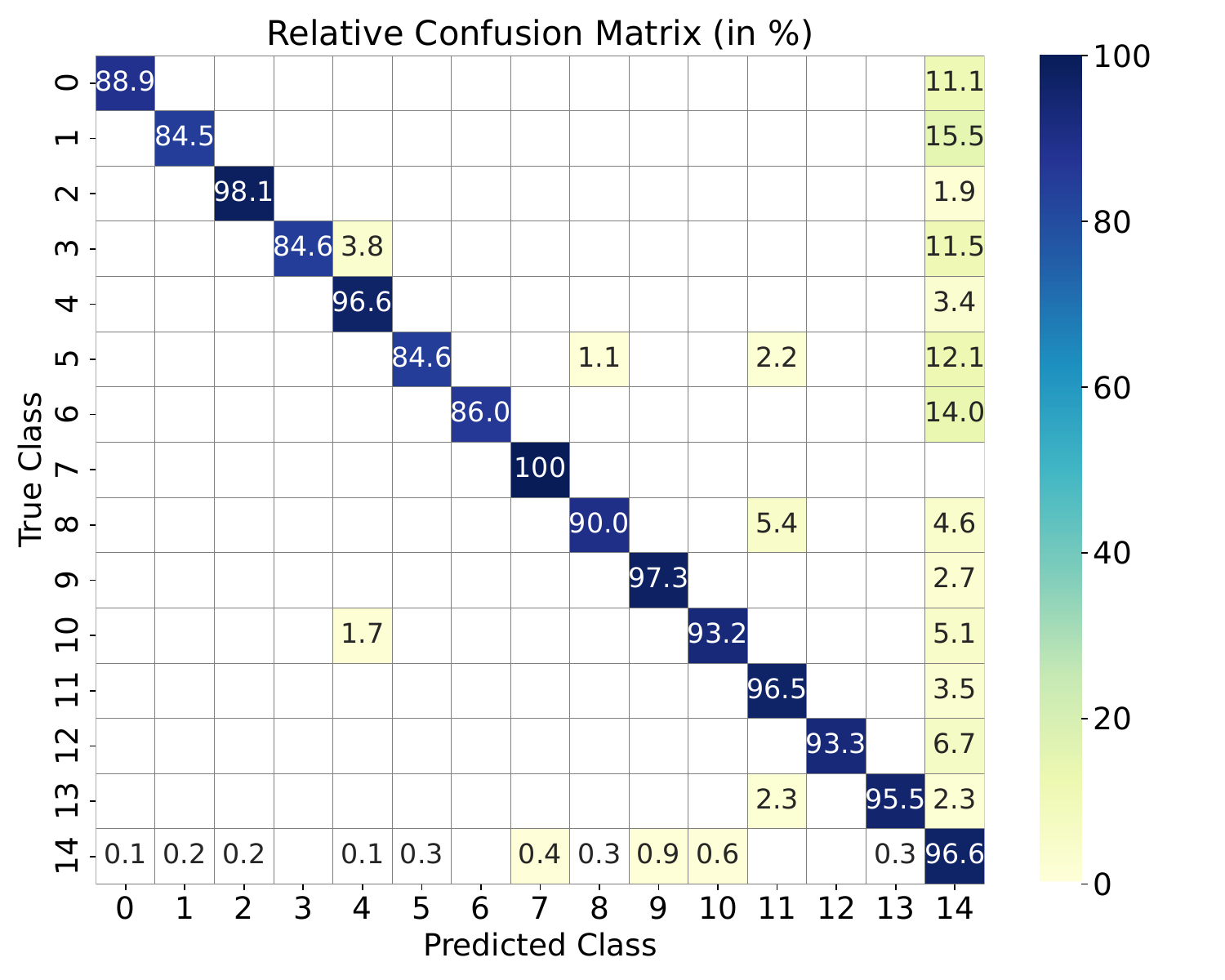}
        \caption{MeTRAbs + HD-GCN}
        \label{fig:conf_hdgcn}
    \end{subfigure}
    \begin{subfigure}[t]{0.32\textwidth}
        \centering
        \includegraphics[width=\linewidth]{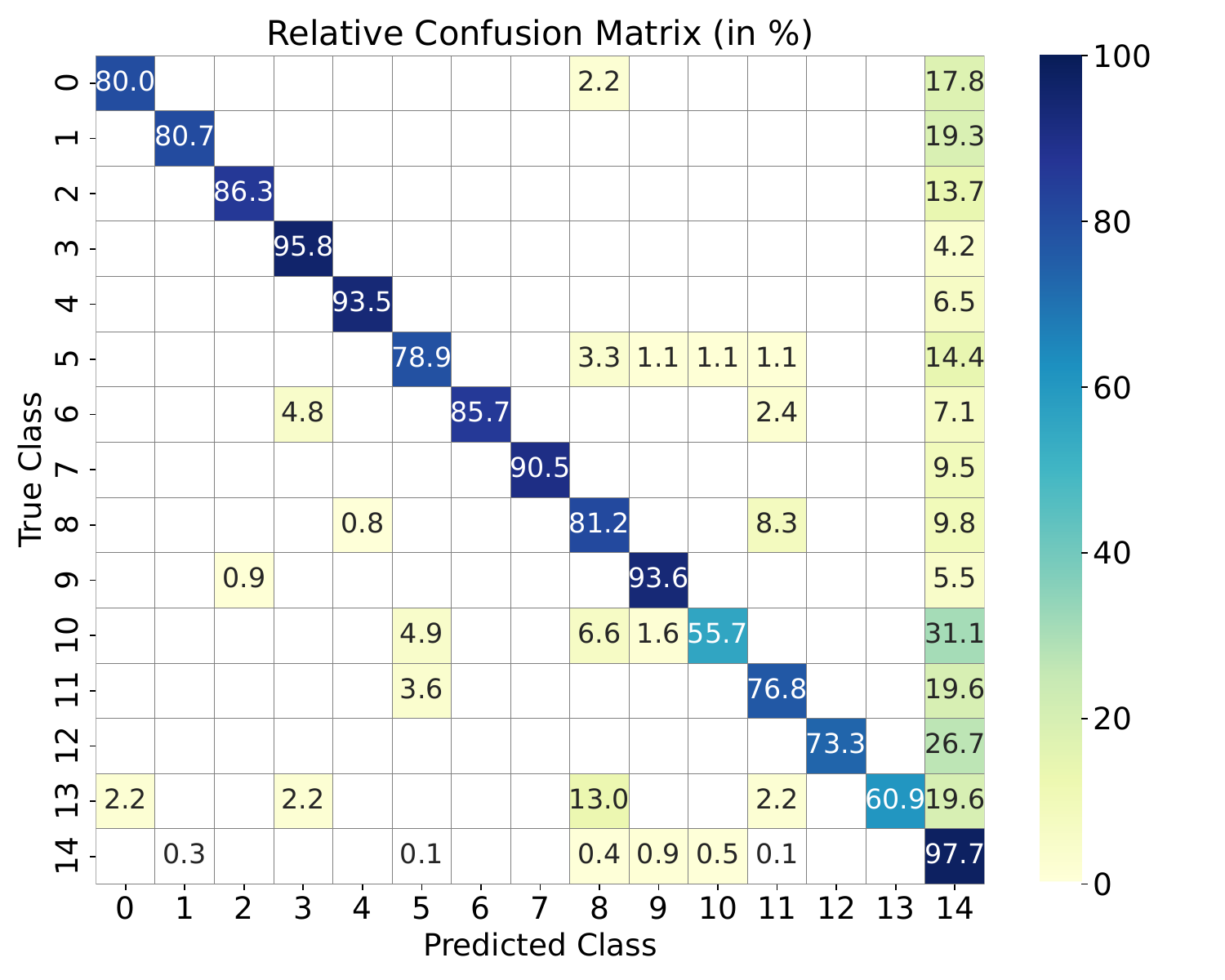}
        \caption{V-JEPA (Attentive Probe)}
        \label{fig:conf_vjepa}
    \end{subfigure}
     \begin{subfigure}[t]{0.32\textwidth}
        \centering
        \includegraphics[width=\linewidth]{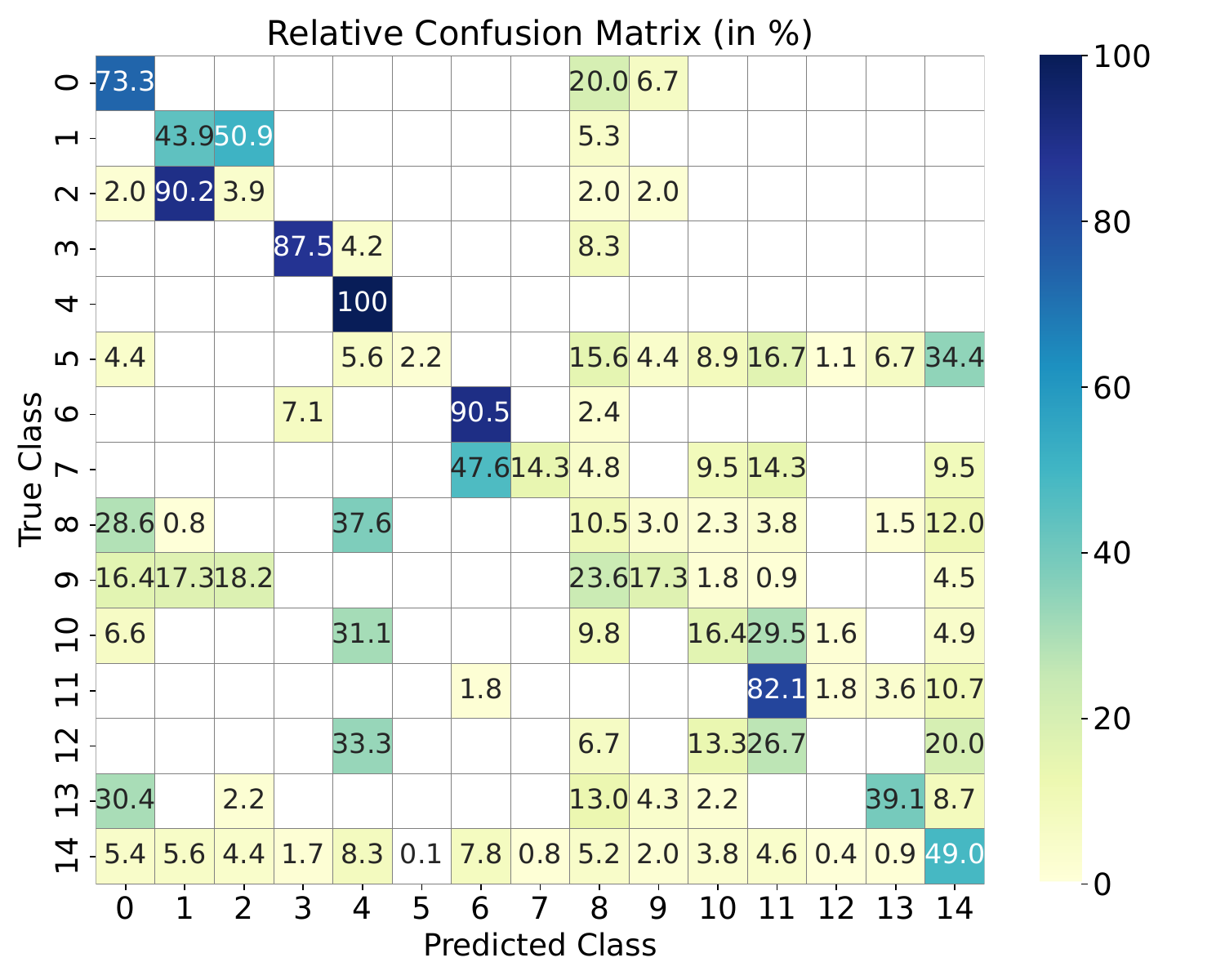}
        \caption{Gemini Flash 2.0 (Video, \textsc{choose})}
        \label{fig:conf_gemini}
    \end{subfigure}
    \caption{Confusion matrices on NUGGET test for MeTRAbs + HD-GCN, V-JEPA (Attentive Probe), and Gemini Flash 2.0 (with video input and \textsc{choose}-prompt). Class IDs correspond to Table~\ref{tab:gestures}. (a) MeTRAbs + HD-GCN achieves high classification quality across all classes with minimal bias towards \textit{Other}. In contrast, V-JEPA (b) exhibits more bias but remains robust for most classes, except for \textit{Slowdown (10)} and \textit{Give me (13)}. Gemini (c) shows significant class confusion.}
    \label{fig:confusion}
\end{figure*}

The confusion matrix (\cref{fig:confusion}a) reveals \textbf{HD-GCN} achieves 100\% class-specific TPR for \textit{Drop off} and $>95\%$ for \textit{Left}, \textit{Follow me}, \textit{Turn back}, \textit{Back}, \textit{Give me}, and \textit{Other}. Lower TPR ($<85\%$) is observed for \textit{Right}, \textit{Terminate}, and \textit{Come}, with a bias towards \textit{Other} due to dataset imbalance. \textbf{V-JEPA} (\cref{fig:confusion}b) reaches 100\% TPR for \textit{Other} and $>95\%$ for \textit{Terminate}. It maintains 85–95\% TPR for \textit{Left}, \textit{Follow me}, \textit{No}, \textit{Drop off}, and \textit{Turn back} but struggles with \textit{Slowdown} (55.7\%) and \textit{Give me} (60.9\%), often misclassifying the latter as \textit{Other} or \textit{Continue}. \textbf{Gemini} (\cref{fig:confusion}c) performs well for \textit{Follow me} (100\%), \textit{No} (90.5\%), and \textit{Terminate} ($>85\%$) but struggles with \textit{Right}, \textit{Give me}, and \textit{Other} ($\sim$40\%).
Severe misclassifications include \textit{Left} vs. \textit{Right} (90.2\% and 50.9\% false negative rate, FNR), \textit{Drop off} vs. \textit{No} (47.6\% FNR), and \textit{Pick up} (100\% FNR), often misclassified as \textit{Follow me}, \textit{Back}, \textit{Other}, and sometimes as \textit{Slowdown} or \textit{Continue}.

\definecolor{lightgray}{gray}{0.9} 

\begin{table}[t]
    \centering
    \renewcommand{\arraystretch}{1.2}
    \setlength{\tabcolsep}{4pt} 
    \caption{Comparison of gesture classification on the NUGGET single-person test set (Jaccard index, $F_1$ score, accuracies). MeTRAbs+HD-GCN outperforms all methods, demonstrating precise gesture recognition. FM Gemini underperforms across all metrics indicating poor gesture understanding, while FM V-JEPA achieves comparable accuracy to HD-GCN but lower Jaccard index and $F_1$ score, suggesting a bias due to class imbalance in the training set.} 
     \label{tab:final_results}
     \vspace*{-2ex}
    \begin{adjustbox}{width=\columnwidth,center}
    \rowcolors{2}{white}{lightgray}
    \begin{tabular}{l l | c c c c}
        \toprule
        & & & & \multicolumn{2}{c}{Acc.$_{\uparrow}$[\unit{\%}] } \\
        \cline{5-6}
        \multirow{-2}{*}{Architecture} &  \multirow{-2}{*}{Backbone} & \multirow{-2}{*}{Jac.$_{\uparrow}$[\unit{\%}]} &  \multirow{-2}{*}{$F_1$$_{\uparrow}$[\unit{\%}]} &  Top-1 & Top-3 \\
        \hline
        \textbf{\acrshort{metrabs}+\acrshort{hdgcn}} & \textbf{EffNetV2-L} & \textbf{87.0} & \textbf{93.0} & \textbf{94.4} & \textbf{100} \\
       
        \acrshort{jepa} (Attentive Pr.) & \acrshort{vit}-L/16 & 77.0 & 86.5 & 90.1 & 97.5  \\
        Gemini (Video, choose) & - &  18.0 & 28.0 &  42.1 & 79.2 \\
   
        \bottomrule
    \end{tabular}
    \end{adjustbox}
    \label{tab:classification_comparison}
    \vspace*{2ex}
\end{table}

\subsection{Discussion}
\textbf{Bias towards \textit{Other} class:} Both HD-GCN and V-JEPA exhibit a bias towards misclassifying gestures as the \textit{Other} class, likely due to the dominance of this class in the NUGGET training data (see \cref{fig:classdist}).

\textbf{Underperformance of zero-shot Gemini:} We find that a zero-shot Gemini-based system could mostly detect user \textit{Attention}/\textit{Termination} commands but struggles heaviliy with directional commands (e.g., \textit{Left}/\textit{Right}) and specific gestures like \textit{Pick up} or \textit{Give me} and thus cannot be used for robot navigation tasks yet. This behavior can be attributed to two main factors: (1) the model is trained for categories, which may not capture fine-grained arm motion and (2) our initial \textit{Other} class definition (random movements), might have been too abstract and may have led to misinterpretations. While more advanced prompting strategies or confidence-based thresholding might address some of these issues, our focus was on testing the limits of zero-shot classification, which is beneficial for non-experts, or when developing task-specific models and collecting training data is prohibitive. 
\section{Ablation Studies}
 \label{Ablation}
To identify the best configuration for our main study, we conducted experiments focusing on (1) comparing linear vs. attentive probing for the V-JEPA classification head, (2) testing various prompting strategies for Gemini, (3) determining the optimal temporal sliding window size for HD-GCN on NUGGET, (3) establishing HD-GCN as robust gesture recognition baseline by comparing it to related methods on another gesture dataset (TCG \cite{data:traffic:wiederer}).

\textbf{Comparing Linear/Attentive Probing on V-JEPA:} \label{jepa_ablation}
As suggested in \cite{Other:V-JEPA:Bardes} we assess attentive and linear probe (last or multi-layer) classification heads for V-JEPA on NUGGET. Attentive Probe outperforms both linear probes, especially in Jaccard (77.0\%) and F1-score (86.5\%), indicating the effectiveness of cross-attention for gesture recognition.

\textbf{Prompting Strategies for Gemini:}  
We use Gemini Flash~2.0 as a zero-shot classifier on NUGGET, where it is prompted to solve a multiple-choice classification problem given visual input. To generate textual descriptions for all gestures, three human annotators independently provided detailed descriptions of gestures from GT videos. We evaluate:

\textit{(a) Granularity Levels:} We use Mistral LLM \cite{other:Mistral7B:Jiang} to merge annotator descriptions and generate 3 varying levels of detail (\cref{tab:gesture_detailed_descriptions}). Merged fine-granular descriptions perform best, though the improvement is small ($<\!\!1\%$).
\textit{(b) Prompting Strategies:} We compare two prompting strategies: including \textit{Other} as a list item ("choose") versus instructing the model to assign \textit{Other} if none of the 14 gestures match.
\textit{(c) Test Data Input Format:} We test mp4 video versus a sequence of 16 frames.
\textit{(d) Bounding Box Crops:} We investigate the effect of using person BBox crops instead of full frames, utilizing BBoxes created by YOLOv8 \cite{other:YOLO_surve:terven} and GT BBoxes. Gemini is not considered for BBox detection, as it performs worse than YOLOv8 (\cref{tab:person_detector_comparison}). We find that Gemini Flash~2.0 performs best with video input and "choose" prompts without prior cropping (\cref{tab:Gemini_Experiments}).
\textit{(e) Gemini Pro~1.5:} Pro~1.5 outperforms Flash~2.0 on the video analysis benchmark VideoMME \cite{other:videomme_fu}. Its best results on NUGGET (42.0\% top-1, video, "choose") are comparable to Flash~2.0, though it requires higher computational costs and runtime \cite{other:Gemini:2025}. Therefore, we exclude it from our main study.

\textbf{Temporal Sliding Window Size:}
\label{slidingwindow} We apply a sliding window instead of pure per-frame processing as it captures temporal dependencies, preventing loss of motion information and reducing gesture misclassifications caused by ambiguous static poses. We tested varying window sizes from 3\,s to 8\,s with 0.5\,s stride using HD-GCN on NUGGET (\cref{fig:own-window-adj}). The optimal 4\,s window achieved 94.4\% accuracy, with highest $F_1$ and recall scores. Thus, we decided to provide annotated 4\,s-sequences for NUGGET. 

\textbf{Testing HD-GCN on TCG:}
To justify HD-GCN, originally developed for action recognition, as a state-of-the-art baseline for \emph{dynamic gesture recognition}, we train and test it on TCG \cite{data:traffic:wiederer}, an existing gesture recognition dataset for autonomous cars. 
In \cref{tab:baselines-eval-best}, we compare HD-GCN to several baseline models (parameter optimized) presented in the TCG paper. All methods were re-trained on TCG using the previously introduced sliding window approach, using the same input 3D poses from Mask R-CNN (2D\,HPE) \cite{heMaskRCNN2018} with 3D uplifting \cite{HPE:videopose3d:pavllo} for a fair comparison. HD-GCN clearly outperforms each baseline by at least 5\% across all metrics in cross-subject and cross-view evaluations, confirming its robustness and competitiveness in gesture recognition.

\definecolor{lightgray}{gray}{0.9}

\begin{table}[t]
    \centering
    \renewcommand{\arraystretch}{1.0}
    \vspace*{2ex}
    \caption{Evaluation results of  MeTRAbs+HD-GCN for different sliding window sizes $\Delta t$  on NUGGET.}
    \label{fig:own-window-adj}
    \vspace*{-2ex}
    \resizebox{\columnwidth}{!}{
    \begin{tabular}{c|c c c c}
    \toprule
    $\Delta t$ [s] & $F_1$$_{\uparrow}$[\unit{\%}] & Prec.$_{\uparrow}$[\unit{\%}] & Rec.$_{\uparrow}$[\unit{\%}] & Acc.$_{\uparrow}$[\unit{\%}] \\
    \midrule
    \rowcolor{lightgray}
    3 & 92.6 & \textbf{94.6} & 91.0 & 94.2 \\
    4 & \textbf{93.0} & 94.1 & \textbf{92.4} & \textbf{94.4} \\
    \rowcolor{lightgray}
    6 & 92.2 & 94.4 & 90.7 & 93.8 \\
    8 & 90.1 & 94.4 & 87.2 & 92.7 \\ 
    \bottomrule
    \end{tabular}
    }
    \vspace*{2ex}
\end{table}

\definecolor{lightgray}{gray}{0.9}

\begin{table}[t!]
    \centering
    \caption{Comparing classification results using V-JEPA with Linear Probing (last layer (LL), multi-layer (ML)) or Attentive Probing heads on  NUGGET test set. Attentive Probing shows superior performance.}
    \label{tab:clsablation}
    \vspace*{-2ex}
    \renewcommand{\arraystretch}{1.2} 
    \setlength{\tabcolsep}{5pt} 
    \rowcolors{2}{white}{lightgray}
    \begin{tabularx}{\linewidth}{l|XXXXX}
    \toprule
     Probing & Jac.$_{\uparrow}$[\unit{\%}] & $F_1$$_{\uparrow}$[\unit{\%}] & Prec.$_{\uparrow}$[\unit{\%}] & Rec.$_{\uparrow}$[\unit{\%}] & Acc.$_{\uparrow}$[\unit{\%}] \\
    \midrule
    Linear (LL) & 25.9 & 37.2 & 57.7 & 34.7 & 68.9 \\
    Linear (ML) & 46.8 & 61.9 & 78.8 & 54.6 & 77.3 \\
    \textbf{Attentive} & \textbf{77.0} & \textbf{86.5} & \textbf{92.8} & \textbf{82.0} & \textbf{90.1} \\
    \bottomrule
\end{tabularx}
\vspace*{2ex}
\end{table}
\begin{table}[t]
    \renewcommand{\arraystretch}{1.1}
    \centering
    \caption{Comparison of Gemini Flash 2.0 input and prompting strategies regarding $F_1$-score and accuracies on our NUGGET test set. Gemini was used zero-shot with the 15-class problem given as multiple choice prompt either with or without if-statement for the \textit{Other} class.}
    \label{tab:Gemini_Experiments}
    \vspace*{-2ex}
    \resizebox{\linewidth}{!}{
        \rowcolors{2}{white}{lightgray}
        \begin{tabular}{l | l c c c c }
            \toprule            
         & & & & \multicolumn{2}{c}{Acc.$_{\uparrow}$[\unit{\%}] } \\
        \cline{5-6}
        \multirow{-2}{*}{Input} & \multirow{-2}{*}{Prompt} & \multirow{-2}{*}{Jac.$_{\uparrow}$[\unit{\%}]} & \multirow{-2}{*}{$F_1$$_{\uparrow}$[\unit{\%}]} &  Top-1 & Top-3 \\
        \hline
            \cellcolor{white} & if & 16.2 & 26.3 & 34.9 & 56.6 \\
            \multirow{-2}{*}{Crop (GT)} & choose & 15.9 & 26.1 & 32.5 & 63.0 \\
            \hline
            
            \cellcolor{white} & if & 16.2 & 26.3 & 34.7 & 56.4 \\
            \multirow{-2}{*}{Crop (YOLO)} & choose & 15.9 & 26.0 & 32.8 & 64.5 \\
            \hline
           
            \cellcolor{white} & if & 15.9 & 25.2 & 34.1 & 56.7 \\
            \multirow{-2}{*}{Frames} & choose & 16.0 & 25.9 & 34.7 & 64.4 \\
            \hline
            
            \cellcolor{white} & if & 17.1 & 26.8 & 39.2 & 68.7 \\
            \multirow{-2}{*}{Video} & \textbf{choose} & \textbf{18.1} & \textbf{28.2} & \textbf{42.1} & \textbf{79.2}
             \\
            \bottomrule
        \end{tabular}
    }
    \vspace*{2ex}
\end{table}

\begin{table}[t]
    \centering
    \caption{YOLOv8 outperforms Gemini Pro 1.5 and Flash 2.0 in person detection quality (IoU = intersection-over-union, as a measure of 2D localization accuracy) on the NUGGET single-person test split.}
    \label{tab:person_detector_comparison}
    \vspace*{-1ex}
    \renewcommand{\arraystretch}{1.1}
    \resizebox{0.7\linewidth}{!}{
    \rowcolors{2}{white}{lightgray}
    \begin{tabular}{l|c c c}
        \toprule   
        Detector & YOLOv8 & Pro 1.5 & Flash 2.0 \\
        \hline
        IoU$_{\uparrow}$[\unit{\%}] & 98.2 & 90.7 & 70.8 \\
        \bottomrule
    \end{tabular}}
    \vspace*{2ex}
\end{table}

\begin{table}[t]
    \centering
    \caption{Examples of textual gesture descriptions with different granularity styles used for prompting Gemini.}
    \vspace*{-2ex}
    \renewcommand{\arraystretch}{1.2} 
    \setlength{\tabcolsep}{5pt} 
    \rowcolors{2}{white}{lightgray} 
    \begin{tabularx}{\linewidth}{l|l|X}
    \toprule
    \textbf{Gesture} & \textbf{Style} & \textbf{Description} \\
    \hline
    Right & Coarse & Pointing to the right \\
    & Normal  & Fully extends their right arm out to their sides \\
    & Detailed & The right arm is fully extended horizontally to the right side of the body, parallel to the ground. The elbow is straight. The shoulder is slightly lowered. \\
    \hline
    No/Stop & Detailed & Both arms are crossed centrally in front of the chest, forming an "X" shape. The hands rest on or near the opposite shoulders or upper arms.  \\
    Follow me & Detailed & Both arms fully extended vertically, positioned directly above the head, indicating an instruction to follow or proceed in a certain direction. \\
    Pick Up & Detailed & A coordinated movement where both arms are lifted symmetrically from a position in front of the body, indicating the gesture of picking up an imaginary object. \\
    \bottomrule
    \end{tabularx}
    \label{tab:gesture_detailed_descriptions}
\end{table}
\definecolor{lightgray}{gray}{0.9}

\begin{table}[t!]
    \centering
    \vspace*{2ex}
    \setlength{\tabcolsep}{4pt}
    \renewcommand{\arraystretch}{1.2}
    \caption{Evaluation of HD-GCN and various other \emph{skeleton-based baselines} from \cite{data:traffic:wiederer} on the TCG dataset following the cross-subject (S) and cross-view (V) evaluation protocols ("Prot.") from \cite{data:traffic:wiederer}. HD-GCN performs best. }
    \label{tab:baselines-eval-best}
    \vspace*{-2ex}
    \begin{adjustbox}{width=\columnwidth,center}
        \rowcolors{2}{white}{lightgray}
        \begin{tabular}{l|cccccc}
            \toprule
            Model & Prot. & Jac.$_{\uparrow}$[\%] & $F_1$$_{\uparrow}$[\%] & Prec.$_{\uparrow}$[\%] & Rec.$_{\uparrow}$[\%] & Acc.$_{\uparrow}$[\%] \\
            \midrule
            
            \cellcolor{white} & S & 76.4 & 85.8 & 84.5 & 88.3 & 83.2 \\
            \multirow{-2}{*}{Bi-GRU} & V & 66.3 & 79.5 & 79.0 & 80.8 & 83.1 \\
            \hline
            
            \cellcolor{white} & S & 41.3 & 56.1 & 54.0 & 76.1 & 69.4 \\
            \multirow{-2}{*}{Bi-LSTM} & V & 59.7 & 74.7 & 71.8 & 82.0 & 76.8 \\
            \hline
            
            \cellcolor{white} & S & 68.9 & 80.0 & 78.9 & 83.4 & 76.2 \\
            \multirow{-2}{*}{GRU} & V & 62.9 & 76.9 & 75.3 & 80.9 & 79.6 \\
            \hline
            
            \cellcolor{white} & S & 45.2 & 61.0 & 57.3 & 69.1 & 66.2 \\
            \multirow{-2}{*}{LSTM} & V & 50.5 & 66.6 & 64.5 & 72.1 & 69.6 \\
            \hline

            \cellcolor{white} & S & 44.2 & 60.3 & 55.7 & 69.4 & 63.8 \\
            \multirow{-2}{*}{RNN} & V & 47.0 & 63.7 & 60.7 & 72.5 & 67.1 \\
            \hline 
            
            \cellcolor{white} & S & \textbf{87.9} & \textbf{93.3} & \textbf{93.4} & \textbf{93.0} & \textbf{93.0} \\
            \multirow{-2}{*}{HD-GCN} & V & \textbf{87.5} & \textbf{93.2} & \textbf{93.4} & \textbf{93.3} & \textbf{95.1} \\

            \bottomrule
        \end{tabular}                       
    \end{adjustbox}
    \vspace*{2ex}
\end{table}

\section{Conclusion}
In this paper, we presented a detailed comparison of an advanced skeleton-based gesture classifier, HD-GCN, with two state-of-the-art FMs, V-JEPA and Gemini 2.0, on our newly proposed NUGGET dataset. NUGGET features intuitive, dynamic upper-body gestures for human-robot communication in intralogistics scenarios across 15 gesture classes.

Our results indicate that the VFM V-JEPA, equipped with an Attentive Probing classification head, demonstrates high robustness, achieving over 90\% accuracy, and shows promise in matching the strong performance of HD-GCN. It is therefore a good candidate for use as a multi-purpose model, which could incorporate additional task-specific heads for robotics downstream tasks such as object detection. However, its lack of language grounding makes it more difficult to use in tasks such as semantic navigation or task planning, which are often specified in natural language.

In contrast, the Gemini VLM, which incorporates language capabilities, significantly underperforms in our prompting experiments. This underscores the limitations of using such a VLM in a zero-shot setting with textual prompts, without domain-specific fine-tuning. Despite the impressive performance of FMs in various benchmarks, we conclude that zero-shot Gemini-based models are not yet suitable for fine-grained gesture recognition tasks involving multiple gesture classes. Our findings suggest that a more effective representation of gestures than current textual descriptions may be needed, or that fine-tuning the VLM on gesture-specific data could enhance performance, as demonstrated by V-JEPA. While graph-based methods like GCNs remain competitive for symbolic gesture recognition, we anticipate that FMs will excel in other downstream tasks where additional scene context is crucial, such as recognizing human-object interactions.

\section*{Acknowledgment}
This work was funded by Robert Bosch GmbH under the project ``Context Understanding for Autonomous Systems'' and the EU Horizon 2020 research and innovation program under grant agreement 101017274 (DARKO).

\AtNextBibliography{\footnotesize}
\printbibliography

\addtolength{\textheight}{-12cm}   

\end{document}